\documentclass[11pt,a4paper]{article}
\usepackage[dvipdfmx]{graphicx}
\usepackage[hyperref]{eacl2021}
\usepackage{times}
\usepackage{latexsym}

\usepackage{microtype}

\aclfinalcopy 

\usepackage{xcolor}

\title{An Investigation Between Schema Linking and Text-to-SQL Performance}

\author{
  Yasufumi Taniguchi\textsuperscript{1} ~
  Hiroki Nakayama\textsuperscript{1} ~
  Kubo Takahiro\textsuperscript{1} ~
  Jun Suzuki\textsuperscript{2,3} \\
  TIS, Inc.\textsuperscript{1} ~~
  Tohoku University\textsuperscript{2} ~~
  RIKEN\textsuperscript{3} \\
  \texttt{\{taniguchi.yasufumi,nakayama.hiroki,kubo.takahiro\}@tis.co.jp} \\
  \texttt{jun.suzuki@ecei.tohoku.ac.jp}
}

\date{}

\begin{document}
\maketitle
\begin{abstract}
Text-to-SQL is a crucial task toward developing methods for understanding natural language by computers.
Recent neural approaches deliver excellent performance; however, models that are difficult to interpret inhibit future developments. Hence, this study aims to provide a better approach toward the interpretation of neural models.
We hypothesize that the internal behavior of models at hand becomes much easier to analyze if we identify the detailed performance of schema linking simultaneously as the additional information of the text-to-SQL performance.
We provide the ground-truth annotation of schema linking information onto the Spider dataset.
We demonstrate the usefulness of the annotated data and how to analyze the current state-of-the-art neural models.\footnote{Our scheme linking annotation on the Spider data is publicly available at: \url{https://github.com/yasufumy/spider-schema-linking-dataset}.
}
\end{abstract}

\section{Introduction}
\label{sec:introduction}
Text-to-SQL is a task to convert the question in natural language to SQL (the logical form).
The attempts to solve text-to-SQL are crucial to establish methodologies for understanding natural language by computers. 
Currently, neural models are widely used for tackling text-to-SQL~\citep{DBLP:journals/corr/abs-2004-03125, zhang-etal-2019-editing, bogin-etal-2019-global, guo-etal-2019-towards,wang-etal-2020-rat}.
However, state-of-the-art neural models on the Spider dataset~\citep{yu-etal-2018-spider}, a current mainstream text-to-SQL benchmark dataset, yield 60--65 exact matching accuracy.
This indicates that current technologies require immense room for improvement to achieve commercialization and utilization as real-world systems.

A severe drawback of the neural approach is the difficulty of analyzing how models capture the clue to solve a task.
Hence, researchers often struggle which direction to focus on to obtain further improvement.
This paper focuses on this problem and considers a methodology that can reduce enormous effort to analyze the model behaviors and find the next direction.
For this goal, we focus on {\it schema linking}.
Schema linking is a special case of {\it entity linking} and a method to link the phrases in a given question with the column names or the table names in the database schema.
\citet{guo-etal-2019-towards} and \citet{wang-etal-2020-rat} show that 
schema linking is an essential module to solve text-to-SQL task effectively.
We hypothesize that if the detailed performance of the schema linking is known simultaneously as additional information for text-to-SQL performance, then the analysis of the internal behavior of the models at hand becomes easier.

To investigate the above-mentioned hypothesis and offer a better analysis of text-to-SQL models, we annotate ground-truth schema linking information onto the Spider dataset~\citep{yu-etal-2018-spider}.
The experiments reveal the usefulness of scheme linking information in the annotated dataset to understand the model behaviors.
We also demonstrate how the current state-of-the-art neural models can be analyzed by comparing the schema linking performance with the text-to-SQL performance.

\section{Related Works}
\paragraph{Text-to-SQL dataset}
There exist many benchmark datasets, such as WikiSQL \citep{zhongSeq2SQL2017}, Adivising \citep{finegan-dollak-etal-2018-improving}, and Spider \citep{yu-etal-2018-spider}. 
WikiSQL is the largest benchmark dataset in text-to-SQL domain.
However, \citet{finegan-dollak-etal-2018-improving} pointed out that WikiSQL includes almost same SQL in the training and test set, because the dataset aims to generate the correct SQL for unknown questions.
They proposed Advising~\citep{finegan-dollak-etal-2018-improving}, which does not include the same SQL in the training and test sets, but it still consists only of SQL with limited clauses from one domain.
\citet{yu-etal-2018-spider} proposed the Spider dataset that includes complicated SQL with many clauses and 138 different domains.
Currently, Spider is considered the most challenging dataset in the text-to-SQL field.

\paragraph{Schema linking}
In text-to-SQL, schema linking is a task to link a phrase in the given question and the table name or the column name.
The methods used for schema linking are often categorized as explicit or implicit approaches.
The explicit approach is treated as the first step of the text-to-SQL pipeline, and thus we obtain the linking information~\citep{yu-etal-2018-typesql,guo-etal-2019-towards,wang-etal-2020-rat}.
In contrast, the implicit approach is a module included in text-to-SQL models, and thus linking is a black box during the process.
To obtain linking information, we mostly focus on the attention module~\citep{DBLP:journals/corr/BahdanauCB14} from question tokens to the database schema mostly equipped by the models in the implicit approach~\citep{krishnamurthy-etal-2017-neural,bogin-etal-2019-representing,bogin-etal-2019-global,zhang-etal-2019-editing,dong-etal-2019-data}.
In this paper, we focus on the explicit approach for a clear discussion.

\section{Scheme Linking Annotation}

\paragraph{Initial dataset}
The Spider dataset~\cite{yu-etal-2018-spider} is a large-scale human annotated and cross-domain text-to-SQL dataset.
The dataset consists of an 8,625 training set, a 1,034 development set, and a 2,147 test set.
Moreover, it contains 200 databases, and no database overlaps in the training, development, and test sets.
We annotate ground-truth schema linking information onto the Spider dataset.
Note that we annotate it only on the development set, not on training and test sets.
This is because this study aims to provide a detailed analysis tool of text-to-SQL models, mainly for investigating the behavior of models and seeking direction for subsequent developments, not to train models for further improving the performance.
Moreover, the test set is not publicly available for the Spider dataset; the test set is only used in the leaderboard system for preventing the \emph{test set tuning} often arose in the evaluation phase.

\begin{figure}[t]
\samepage
\centering
\includegraphics[scale=0.25]{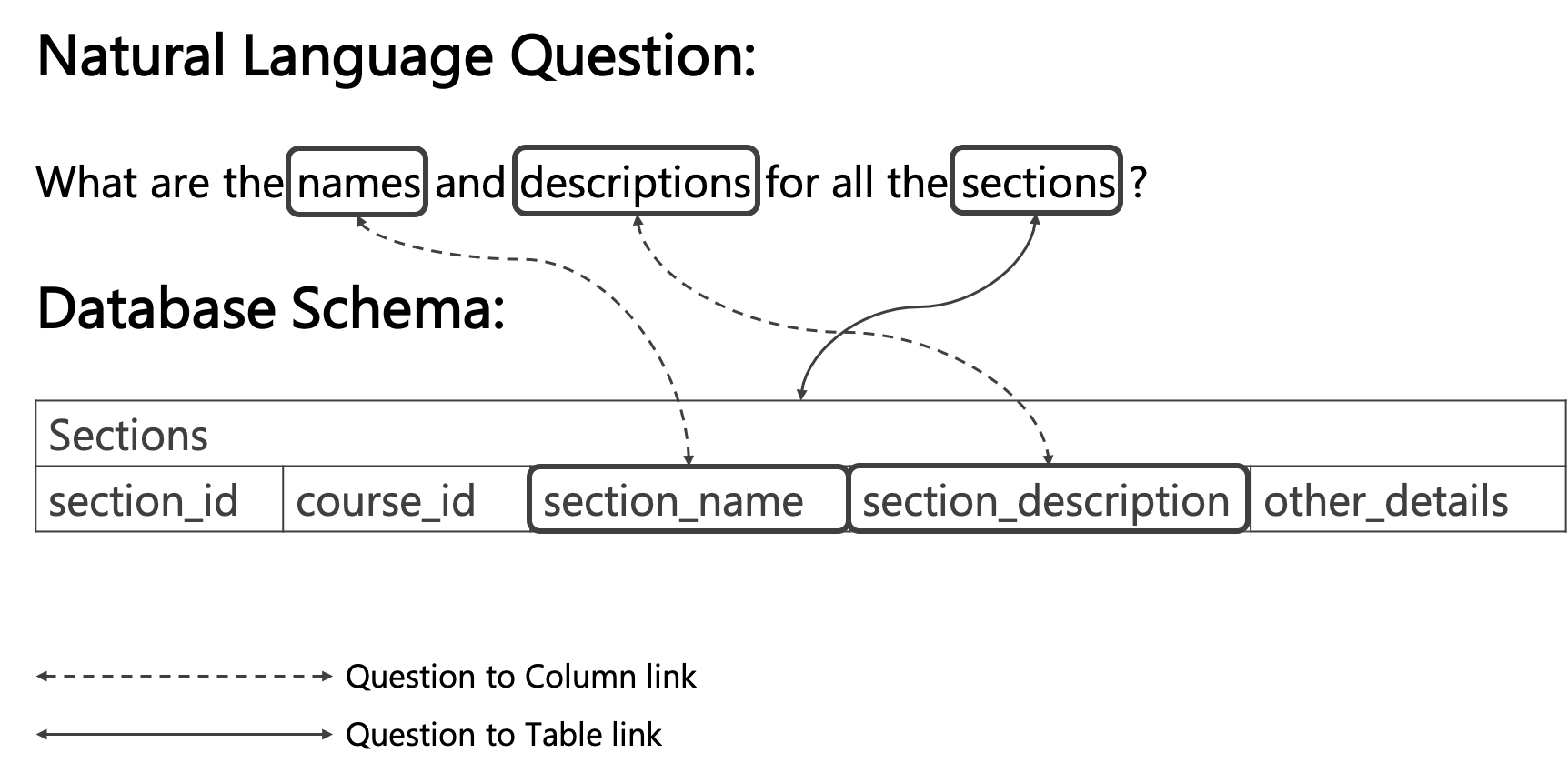}
\caption{Annotated example}
\label{fig:annotation-example}
\end{figure}
\begin{table}[t]
\centering
\small
\tabcolsep 2.8pt
 \begin{tabular}{l|ccccc}
                    & \#LABEL  & MAX & MIN & AVG  & STD \\
 \hline
 Total              & 3,077 & 8   & 1   & 2.98 & 1.227 \\
 \,\,Table              & 1,223 & 5   & 0   & 1.18 & 0.751 \\
 \,\,Column             & 1,854 & 0   & 0   & 1.79 & 0.997 \\
 Total ($l=1$)      & 2,359 & 8   & 0   & 2.28 & 1.229 \\
 \,\,Table ($l=1$)      & 1,031 & 5   & 0   & 1.00 & 0.764 \\
 \,\,Column ($l=1$)     & 1,328 & 0   & 0   & 1.28 & 0.948 \\
 Total ($l \ge 2$)  & 718   & 4   & 0   & 0.69 & 0.851 \\
 \,\,Table ($l \ge 2$)  & 192   & 3   & 0   & 0.19 & 0.424 \\
 \,\,Column ($l \ge 2$) & 526   & 0   & 0   & 0.51 & 0.751 \\
 \end{tabular}
\caption{Statistics of the annotated data for each sentence. \#LABEL: number of label for a sentence, MAX: the maximum number of labels, MIN: the minimum number of labels, AVG: the average number of labels, STD: the standard deviation for the number of labels.}
  \label{tb:anno-stats}
\end{table}

%
\paragraph{Annotation detail and statistics}
The annotation is performed  by two software engineers who are familiar with SQL.
They use Doccano\footnote{https://github.com/doccano/doccano} as the annotation tool.
Figure~\ref{fig:annotation-example} shows an annotation example\footnote{See several other examples in Appendix~\ref{sec:anno_example}}.
Table~\ref{tb:anno-stats} shows the statistics of the annotated data.

\paragraph{Quality check}
For the annotation quality check, we validate the annotation agreement between two annotators by independently annotating the same 100 examples.
The annotation agreement of Cohen's kappa is 0.764 ($95\% \mbox{ CI}= 0.722 - 0.806$, $p<0.01$)\footnote{
The un-annotated tokens unfairly increase the kappa score on sequence segmentation tasks~\citep{brandsen-etal-2020-creating}.
We follow the instruction written in ~\citet{brandsen-etal-2020-creating} to calculate the kappa score only on tokens, either one annotated.
}.
According to \citet{Landis77}, the kappa value in the range $0.61-0.80$ is categorized in \emph{substantial agreement}.
Moreover, the ${\rm F}_1$ score of annotation of two annotators is 87.5.
We calculate the ${\rm F}_1$ score as suggested in several previous studies~\cite{brandsen-etal-2020-creating,grouin-etal-2011-proposal,alex-etal-2010-agile}.
According to these results, we believe that our annotated scheme linking data are highly reliable as the ground truth.

\paragraph{Data split}
We split the annotated data into two distinct sets and used one for the development set and another for the test set.
Hereafter, we refer to these new sets as the \emph{development set} and the \emph{test set}, respectively; it is crucial to note that this paper does not deal with the true test data in the Spider dataset.
Consequently, the development and test sets both contain 517 examples for each\footnote{We confirmed that there is no database overlapping between the new development and test sets.This is the identical configuration for the original Spider dataset.}.

\paragraph{Evaluation metric}
Schema linking is a task similar to the named entity recognition and relation extraction~\cite{marsh-perzanowski-1998-muc}.
Therefore, we calculate the precision, recall, and ${\rm F}_1$-score~\cite{tjong-kim-sang-2002-introduction,tjong-kim-sang-de-meulder-2003-introduction} for evaluating the schema linking performance.

\section{Experiment}
\begin{table}
\centering
\small
\tabcolsep 2.4pt
\begin{tabular}{l|c|l}
\textbf{name} & \textbf{alias} & \textbf{explanation}\\ \hline
w/o uni-gram-(a)  & a & The uni-grams are ignored \\ \hline
w/o uni-gram-(b)  & b & \begin{tabular}[x]{@{}l@{}}The uni-grams are ignored. \\The partial matches are ignored\end{tabular} \\  \hline
w/o column-match  & c & The column names are ignored \\ \hline
w/o table-match   & d & The table names are ignored  \\ \hline
only uni-gram-(a) & e & \begin{tabular}[x]{@{}l@{}}Only the uni-grams \\ are considered. \end{tabular}\\  \hline
only uni-gram-(b) & f & \begin{tabular}[x]{@{}l@{}}Only the uni-grams \\ are considered. \\ However, the partial \\ matches are ignored. \end{tabular} \\  \hline
random            & g & Randomly linking\\  \hline
w/o all           & h & No schema linking \\ 
\end{tabular}
\caption{\label{methods-table} Schema linking methods. We use the alias in later experiments.}
\end{table}
\begin{table}[t]
\centering
\small
\tabcolsep 2.0pt
 \begin{tabular}{l|r|rrrrrr}
 & \multicolumn{1}{c|}{Spider } 
 & \multicolumn{6}{c}{Schema Linking} \\
 Model 
 & \multicolumn{1}{c|}{EM} 
 & \multicolumn{1}{c}{${\rm F}_1$ }
 & \multicolumn{1}{c}{Pre.}
 & \multicolumn{1}{c}{Rec.} 
 & \multicolumn{1}{c}{\#FP} 
 & \multicolumn{1}{c}{\#FN} 
 & \multicolumn{1}{c}{\#TP} \\
 \hline
 IRNet   & 58.8 & 72.6 & 79.9 & 66.5 & 243 & 487 & 967 \\
 RAT-SQL & 69.2 & 58.1 & 46.0 & 78.8 & 1,345 & 308 & 1,146 \\
 \end{tabular}
\caption{Schema linking and text-to-SQL results. EM: exact match, Pre.: precision, Rec.:recall, \#FP: number of false positive, \#FN: number of false negative, \#TP: number of true positive.}
  \label{tb:main_result}
\end{table}
\begin{table}[t]
\centering
\small
\tabcolsep 1.8pt
 \begin{tabular}{l|r|rrrrrr}
 & \multicolumn{1}{c|}{Spider } 
 & \multicolumn{6}{c}{Schema Linking} \\
 Model 
 & \multicolumn{1}{c|}{EM} 
 & \multicolumn{1}{c}{${\rm F}_1$ }
 & \multicolumn{1}{c}{Pre.}
 & \multicolumn{1}{c}{Rec.} 
 & \multicolumn{1}{c}{\#FP} 
 & \multicolumn{1}{c}{\#FN} 
 & \multicolumn{1}{c}{\#TP} \\
 \hline
 IRNet    & 58.8  & 72.6  & 79.9 & 66.5 & 243 & 487 & 967 \\
 IRNet-a  & 55.3  & 54.9  & 79.4 & 42.0 & 158 & 844 & 610 \\
 IRNet-b  & 56.7  & 54.3  & 79.3 & 41.3 & 157 & 854 & 600 \\
 IRNet-c  & 53.8  & 49.2  & 78.0 & 36.0 & 148 & 930 & 524 \\
 IRNet-e  & 53.0  & 48.5  & 70.7 & 36.9 & 223 & 917 & 537 \\
 IRNet-d  & 51.8  & 47.5  & 70.3 & 35.8 & 220 & 933 & 521 \\
 IRNet-f  & 50.9  & 39.1  & 71.8 & 26.8 & 153 & 1,064 & 390 \\
 IRNet-g  & 47.8  & 24.3  & 28.5 & 21.1 & 769 & 1,147 & 307 \\
 IRNet-h  & 49.0  &  0.0  & 0.0  & 0.0  & 0   & 1,454 & 0 \\
 \hline\hline
 RAT-SQL   & 69.2 & 58.1 & 46.0 & 78.8 & 1,345 & 308   & 1,146  \\
 RAT-SQL-f & 62.3 & 67.3 & 76.3 & 60.2 & 272   & 578   & 876    \\
 RAT-SQL-b & 42.9 & 17.5 & 34.4 & 11.7 & 324   & 1,284 & 170    \\
 RAT-SQL-g & 58.8 & 22.0 & 33.1 & 16.5 & 486   & 1,214 & 240 \\
 RAT-SQL-h & 58.2 & 0.0    & 0.0    & 0.0    & 0     & 1,454 & 0 \\

 \end{tabular}
\caption{Schema linking and text-to-SQL results. 
}
  \label{tb:sl-and-sql}
\end{table}
\begin{table}[t]
\centering
\small
\tabcolsep 1.5pt
  \begin{tabular}{l|r|rrrrrr}
 & \multicolumn{1}{c|}{Spider } 
 & \multicolumn{6}{c}{Schema Linking} \\
 Model 
 & \multicolumn{1}{c|}{EM} 
 & \multicolumn{1}{c}{${\rm F}_1$ }
 & \multicolumn{1}{c}{Pre.}
 & \multicolumn{1}{c}{Rec.} 
 & \multicolumn{1}{c}{\#FP} 
 & \multicolumn{1}{c}{\#FN} 
 & \multicolumn{1}{c}{\#TP} \\
 \hline
 IRNet anno & 62.5 & 100.0    & 100.0    & 100.0    & 0   & 0   & 1,454  \\
 IRNet mix   & 59.2 & 78.3 & 85.7 & 72.0 & 175 & 407 & 1,047  \\
 \hline
 \hline
 RAT-SQL anno & 69.6 & 100.0    & 100.0    & 100.0    & 0   & 0   & 1,454  \\
 RAT-SQL mix & 69.1 & 79.2 & 71.9 & 88.3 & 503 & 170 & 1,284  \\
 \end{tabular}
\caption{Evaluation on annotated dataset (anno) and mixing the annotations and estimated predictions (mix). 
}
  \label{tb:eval-on-annotation}
\end{table}
This section demonstrates the utilization of the proposed annotated dataset to understand the model behavior and determine the next directions for further improvement.

\subsection{Experiment Setup}

\paragraph{Evaluation metric}ed the Spider dataset to evaluate the Text-to-SQL performance.
We used the exact matching (\textbf{EM}) accuracy for the evaluation of text-to-SQL performance of the Spider dataset as introduced in \citet{yu-etal-2018-spider}%
\footnote{We used the official evaluation script provided by \citet{yu-etal-2018-spider}}.
Moreover, we evaluated the schema linking performance by F$_1$-score, as explained in the previous section.

\paragraph{Baseline models}
We selected IRNet \citep{guo-etal-2019-towards} and RAT-SQL \citep{wang-etal-2020-rat} for the baseline models of the experiments to reveal the effectiveness of the proposed dataset, where we refer to them as \texttt{IRNet} and \texttt{RAT-SQL}, respectively\footnote{See \ref{seq:model} for the detailed descriptions of \texttt{IRNet} and \texttt{RAT-SQL}.}.
It should be noted that both of their models employed an explicit approach, whose first steps are scheme linking; thus, their settings match to evaluate the usefulness of the proposed annotated data. 
However, we also emphasize here that their schema linking methods differ from each other although their methods consist of combinations of similar multiple rules, where \texttt{IRNet} maps the phrase to the single table or column, and \texttt{RAT-SQL} maps the phrase to the multiple tables or columns.
Further, \texttt{IRNet} and \texttt{RAT-SQL} mark the the top-line scores in the leader board of the Spider dataset\footnote{https://yale-lily.github.io/spider}; specifically,  \texttt{RAT-SQL} is the current state-of-the-art model.
These facts suggest to use them as baseline models in our experiments.
We selected the identical hyper-parameter values for both \texttt{IRNet} and \texttt{RAT-SQL} with their original papers, i.e., \citet{guo-etal-2019-towards} and  \citet{wang-etal-2020-rat}.

\paragraph{Investigations}
The schema linking methods used in \texttt{IRNet} and \texttt{RAT-SQL} follow the rule-based approach\footnote{See Appendix~\ref{sec:scheme_linking} for the rules used in their methods.} that allows easier interpretation of model behavior.
To investigate the usefulness of the schema linking information, we conducted the fine-grained schema linking ablation experiments.
Through these experiments, we explore the general behaviors of text-to-SQL models when the performance of scheme linking changes.
To accomplish this, we prepare eight methods shown in Table~\ref{methods-table}.
%

\section{Results and analysis}
\paragraph{Behaviors of \texttt{IRNet} and \texttt{RAT-SQL}}
Table~\ref{tb:main_result} shows the results of the schema linking and text-to-SQL performance of \texttt{IRNet} and \texttt{RAT-SQL}. 
The Spider EM of \texttt{RAT-SQL} is significantly better than that of \texttt{IRNet}, whereas the scheme linking F$_{1}$ of \texttt{RAT-SQL} is much worse than that of \texttt{IRNet}.
This mismatch occurred by the difference of the scheme linking strategy as  \texttt{RAT-SQL} prioritizes recall over precision, as presented in Table~\ref{tb:main_result}.

\paragraph{Correlation}
Table~\ref{tb:sl-and-sql} shows a type of ablation study to gradually decrease the F$_{1}$ scores by eliminating the schema linking rules.
Further, Table~\ref{tb:eval-on-annotation} shows the simulated evaluation results when we obtained the perfect prediction (F$_1=100$), or better predictions than that of the original \texttt{IRNet} and \texttt{RAT-SQL}\footnote{We obtain "anno" from the human annotation, and "mix" by randomly choosing the example from the human annotation or the original schema linking result}.
We observed a strong correlation between scheme linking F$_{1}$ and Spider EM on \texttt{IRNet}.
In fact, the correlation coefficient between them is 0.937 with $p=2.7\times10^{-5}$.
This fact indicates that the scheme linking considerably affects the final Spider EM score. 
Thus, we can roughly estimate the EM scores from scheme linking F$_{1}$ without performing the entire training and evaluation procedures of \texttt{IRNet}.
Unlike \texttt{IRNet}, the Spider EM of \texttt{RAT-SQL} seems not to be strongly correlated to the scheme linking F$_1$.
The correlation coefficient between them is 0.737 with $p=0.058$.
However, if we checked the Spider EM and \#TP correlation as \texttt{RAT-SQL} prioritizes recall than precision, it becomes 0.81 with $p=0.027$.
Therefore, \texttt{RAT-SQL} still has a strong correlation between scheme linking results.
Additionally, the Spider EM for \texttt{IRNet anno} is higher than the original \texttt{IRNet} (62.5 vs. 58.8).
Similarly, \texttt{RAT-SQL anno} is higher than for the original \texttt{RAT-SQL} (69.6 vs. 69.2).
These results also support the reliability of the proposed annotation as the performance gain should be derived from the correct (better) scheme linking.

\begin{figure}[t]
\centering
\tabcolsep 2pt
\footnotesize
 \begin{tabular}{c|l}
 
 Question & How many countries exist? \\
 \hline
 Gold     & \begin{minipage}{5cm}\vspace{0.25cm}\begin{verbatim}
 SELECT count(*)
 FROM   COUNTRIES;\end{verbatim}\vspace{0.15cm}\end{minipage} \\
 \hline\hline
 
 IRNet-f  & \begin{minipage}{5cm}\vspace{0.25cm}\begin{verbatim}
 SELECT count(*)
 FROM   COUNTRIES;\end{verbatim}\vspace{0.15cm}\end{minipage} \\
 \hline
 
 IRNet-h  & \begin{minipage}{5cm}\vspace{0.25cm}\begin{verbatim}
 SELECT count(T1.Country)
 FROM   car_makers AS T1;\end{verbatim}\end{minipage}
 
 \end{tabular}
  \caption{Example of IRNet output.}
  \label{fig:irnet-output-example}

\end{figure}

\paragraph{Error analysis}
Figure~\ref{fig:irnet-output-example} shows actual examples of \texttt{IRNet} outputs\footnote{The actual examples obtained from \texttt{RAT-SQL} are presented in Appendex~\ref{sec:output} because of the space limitation.}.
IRNet-f successfully generates the correct SQL query, while IRNet-h does not.
In the absence of scheme linking annotation, it is relatively difficult to determine the cause of this failure.
However, using the scheme linking annotation, 
we can easily find the reason for the failure of IRNet-h; it failed to link {\it countries} in the question as to the {\it table name}.
This is a simple example of leveraging the proposed annotation for analyzing the model behaviors.
We believe there are many ways to utilize the proposed annotation to further analyze the model behaviors.

\section{Conclusion}
The schema linking is an essential module for performing the text-to-SQL task effectively.
We annotated the schema linking information onto the Spider dataset.
Then, we investigated the usefulness of the proposed annotation to understand the model behaviors of text-to-SQL models and seek the next directions for further development.

As a demonstration, we selected IRNet and RAT-SQL, which are the state-of-the-art methods on the Spider data, and evaluated both scheme linking and Spider EM scores.
The results showed strong correlations between the schema linking F$_1$ and Spider EM scores for IRNet and the number of true positive and Spider EM scores for RAT-SQL.
These correlations may offer a rough estimation of the final Spider exact match scores without training the models.
We hope the proposed scheme linking annotation helps future studies in the text-to-SQL task.

\bibliography{anthology,eacl2021}
\bibliographystyle{acl_natbib}

\clearpage

\appendix

\section{Spider data}

\begin{figure}[ht]
\centering
\tabcolsep 2pt
\footnotesize
 \begin{tabular}{c|l}
 Question & \begin{tabular}{@{}l@{}}What are the names and \\
 the descriptions for all the sections? 
 \end{tabular} \\
 \hline
 SQL & \begin{minipage}{5cm}
 \vspace{0.25cm}
 \begin{verbatim}
 SELECT section_name,
         section_description
 FROM   Sections;
 \end{verbatim}
 \end{minipage}
 \end{tabular}

  \caption{Example pair of the question and SQL query of Spider  \citep{yu-etal-2018-spider}.
}
  \label{fig:text-to-sql-example}

\end{figure}

Figure~\ref{fig:text-to-sql-example} shows an example in the dataset. 
A single data sample is constructed by the natural language question and the SQL query.

\section{Details of scheme linking methods}
\label{sec:scheme_linking}
The scheme linking methods in IRNet and RAT-SQL both classify the word n-grams in questions to three classes, namely, {\it table}, {\it column}, or {\it NONE}.
Then, they enumerate the word n-grams of length 1-6 in the question and classify longer n-grams first.
During the scan of the n-grams, it classifies {\it column} or {\it table} when the n-gram matches exactly or partially.
If the n-gram matches both {\it column} and {\it table}, {\it column} is prioritized.
If the n-gram matches nothing, that n-gram is classified to {\it NONE}

\section{Details of baseline models}
\label{seq:model}
IRNet is the model that successfully utilizes schema linking.
IRNet has the three stages to generate the SQL query.
The first stage is the schema linking explained above.
The second stage is the main part of this model.
It consists of generation of SemQL, which is the immediate representation between the question and SQL query.
SemQL has a much simpler grammar than SQL.
The last stage converts SemQL to SQL.

RAT-SQL is the first-place model on the Spider leader board.
RAT-SQL also uses the schema linking technique proposed in IRNet.
In RAT-SQL, \citet{wang-etal-2020-rat} proposed the {\it relation-aware self-attention}, which effectively encodes the directed graph of the database schema.
Their approach uses self attention mechanism \citep{DBLP:journals/corr/VaswaniSPUJGKP17} to combine the phrases in the database schema and the phrases in the question.

\section{Output examples}
\label{sec:output}

We show the \texttt{RAT-SQL} outputs in Figure~\ref{fig:ratsql-output-example}.
From Figure \ref{fig:ratsql-output-example}, both models fail to generate the SQL query.
However, RAT-SQL successfully predicates the SELECT clauses, while RAT-SQL-f does not.
This is because the schema linking of RAT-SQL can capture the bi-gram matches.

\begin{figure}[t]
\centering
\tabcolsep 2pt
\footnotesize
 \begin{tabular}{c|l}
 
 Question & \begin{tabular}{@{}l@{}}Find the first name, country code and birth date \\ 
 of the winner who has the highest rank points \\
 in all matches.  
 \end{tabular} \\
 \hline
 Gold & \begin{minipage}{5cm}\vspace{0.25cm}\begin{verbatim}
 SELECT T1.first_name, 
        T1.country_code, 
        T1.birth_date 
 FROM   players AS T1 
        JOIN matches AS T2 
          ON T1.player_id = 
             T2.winner_id 
 ORDER  BY 
 T2.winner_rank_points DESC 
 LIMIT  1  \end{verbatim}\vspace{0.2cm}\end{minipage} \\
 \hline\hline
 
 RAT-SQL & \begin{minipage}{5cm}\vspace{0.25cm}\begin{verbatim}
 SELECT players.first_name, 
        players.country_code, 
        players.birth_date 
 FROM   players 
        JOIN matches 
          ON players.player_id 
             = 
             matches.loser_id 
 ORDER  BY matches.winner_ht 
           ASC 
 LIMIT  1\end{verbatim}\vspace{0.2cm}\end{minipage} \\
 \hline
 
 RAT-SQL-f & \begin{minipage}{5cm}\vspace{0.25cm}\begin{verbatim}
 SELECT players.first_name, 
        rankings.ranking_date, 
        matches.tourney_date 
 FROM   players 
        JOIN matches 
          ON players.player_id 
             = 
             matches.loser_id 
        JOIN rankings 
          ON players.player_id 
             = 
             rankings.player_id 
 ORDER  BY rankings.ranking ASC 
 LIMIT  1  \end{verbatim}\vspace{0.2cm}\end{minipage}
 
 \end{tabular}

  \caption{Example of RAT-SQL output.}
  \label{fig:ratsql-output-example}

\end{figure}

\section{Annotated dataset examples}
\label{sec:anno_example}
We show our annotated dataset examples randomly picked from Figure~\ref{fig:real-anno-data}.

\begin{figure*}[ht]
\begin{center}
\small
\begin{verbatim}
{
  "question": "Count the number of templates.",
  "labels": [[20, 29, "Templates"]]
}
{
  "question": "Which airline has abbreviation 'UAL'?",
  "labels": [[6, 13, "airlines.Airline"], [18, 30, "airlines.Abbreviation"]]
}
{
  "question": "Show the names of high schoolers who have likes, and numbers 
  of likes for each.",
  "labels": [[9, 14, "Highschooler.name"], [18, 32, "Highschooler"], 
             [42, 47, "Likes"], [64, 69, "Likes"]]
}
{
  "question": "How many orchestras does each record company manage?",
  "labels": [[9, 19, "orchestra"], [30, 44, "orchestra.Record_Company"]]
}
{
  "question": "What is the first name of every student who has a dog but 
  does not have a cat?",
  "labels": [[12, 22, "Student.Fname"], [32, 39, "Student"]]
}
{
  "question": "Show different citizenship of singers and the number of 
  singers of each citizenship.",
  "labels": [[15, 26, "singer.Citizenship"], [30, 37, "singer"], 
             [56, 63, "singer"], [72, 83, "singer.Citizenship"]]
}
{
  "question": "What are 3 most highly rated episodes in the TV series 
  table and what were those ratings?",
  "labels": [[23, 28, "TV_series.Rating"], [29, 37, "TV_series.Episode"], 
             [45, 54, "TV_series"], [81, 88, "TV_series.Rating"]]
}
{
  "question": "Find the semester when both Master students and Bachelor 
  students got enrolled in.",
  "labels": [[9, 17, "Student_Enrolment.semester_id"], 
             [35, 43, "Degree_Programs.degree_summary_name"],
             [57, 65, "Degree_Programs.degree_summary_name"], 
             [70, 81, "Student_Enrolment"]]
}
{
  "question": "What are the contestant numbers and names of the 
  contestants who had at least two votes?",
  "labels": [[13, 23, "CONTESTANTS"], 
             [24, 31, "CONTESTANTS.contestant_number"],
             [36, 41, "CONTESTANTS.contestant_name"],
             [49, 60, "CONTESTANTS"], [82, 87, "VOTES"]]
}
{
  "question": "Show names, results and bulgarian commanders of the 
  battles with no ships lost in the 'English Channel'.",
  "labels": [[5, 10, "battle.name"], [12, 19, "battle.result"], 
             [24, 44, "battle.bulgarian_commander"], [52, 59, "battle"],
             [68, 73, "ship"], [74, 78, "ship.lost_in_battle"], 
             [79, 81, "ship.location"]]
}
\end{verbatim}
\end{center}
\caption{Example of our annotated data. The {\it labels} field provides the annotation information}
\label{fig:real-anno-data}
\end{figure*}

\end{document}